%% file: main.tex
\documentclass[10pt,twocolumn,letterpaper]{article}

\usepackage{iccv}              


\definecolor{iccvblue}{rgb}{0.21,0.49,0.74}
\usepackage[accsupp]{axessibility}
\usepackage[pagebackref,breaklinks,colorlinks,allcolors=iccvblue]{hyperref}
\usepackage{bm}
\usepackage{xcolor}
\usepackage{comment}
\usepackage{booktabs}
\def\paperID{3169} 
\def\confName{ICCV}
\def\confYear{2025}

\title{OCSplats: Observation Completeness Quantification and Label Noise Separation in 3DGS}

\author{Han Ling$^{1}$, Xian Xu$^{2,3}$, Yinghui Sun$^{2}$, Quansen Sun$^{1}$\thanks{Corresponding author}\\
$^{1}$Nanjing University of Science and Technology\quad $^{2}$ Southeast University\quad $^{3}$CityU HK\\
{\tt\small 1536233573@qq.com,230248289@seu.edu.cn}
}

\begin{document}
\maketitle
\input{sec/0_abstract}    
\input{sec/1_intro}
\input{sec/2_relateworks}
\input{sec/3_method}
\input{sec/4_exp}

{
    \small
    \bibliographystyle{ieeenat_fullname}
    \bibliography{main}
}

\input{sec/X_suppl}

\end{document}

%% file: sec/0_abstract.tex
\begin{abstract}
3D Gaussian Splatting (3DGS) has become one of the most promising 3D reconstruction technologies. However, label noise in real-world scenarios—such as moving objects, non-Lambertian surfaces, and shadows—often leads to reconstruction errors. Existing 3DGS-Bsed anti-noise reconstruction methods either fail to separate noise effectively or require scene-specific fine-tuning of hyperparameters, making them difficult to apply in practice.
This paper re-examines the problem of anti-noise reconstruction from the perspective of epistemic uncertainty, proposing a novel framework, OCSplats. By combining key technologies such as hybrid noise assessment and observation-based cognitive correction, the accuracy of noise classification in areas with cognitive differences has been significantly improved.
Moreover, to address the issue of varying noise proportions in different scenarios, we have designed a label noise classification pipeline based on dynamic anchor points. This pipeline enables OCSplats to be applied simultaneously to scenarios with vastly different noise proportions without adjusting parameters. 
Extensive experiments demonstrate that OCSplats always achieve leading reconstruction performance and precise label noise classification in scenes of different complexity levels. Code is available \footnote{\url{github.com/HanLingsgjk/OCSplats}}.
\end{abstract}

%% file: sec/1_intro.tex
\section{Introduction}
The new view synthesis addresses the challenge of previously unseen viewpoints. 3D Gaussian Splatting~\cite{kerbl20233d,yu2024gsdf,qian20243dgs} has recently emerged as a groundbreaking technique in this domain, primarily due to its ability to reconstruct geometrically consistent point cloud scenes rapidly. As a result, it has found broad applications in areas such as depth estimation~\cite{ling2024sag,tosi2023nerf}, optical flow training~\cite{Ling_2024_CVPR,ling2022scale}, and autonomous driving~\cite{zhou2024drivinggaussian,han2024ggs,hess2024splatad,li2024vdg,ling2023learning}.

However, training a 3DGS model requires static scene images with known camera poses. In real-world scenarios, non-Lambertian~\cite{bakshi1994shape} surfaces, moving objects, and shadows introduce noise into the labels for 3DGS training. Eliminating such noise is nontrivial. Beyond apparent moving objects~\cite{sabour2023robustnerf,ren2024nerf}, more subtle phenomena—like shadows and reflections on non-Lambertian surfaces—demand manual per-pixel annotation. These factors underscore the limitations of current reconstruction methods when applied to complex real-world scenes.

\begin{figure}[!t]
	\centering
	\includegraphics[width=3.0in]{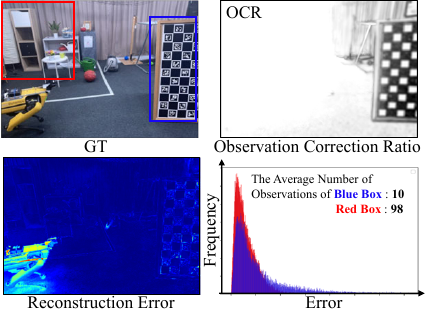}
	\caption{\textbf{Observation Cognitive Correction.} The error histogram in the bottom right indicates that even without occlusion, the reconstruction error of the missing observation areas (blue box) in the photo group is often greater than that of the areas with more observations (red box). We proposed the observation cognitive correction technique to measure  these difficult areas (blue boxes) and corrected the noise assessment through OCR when classifying label noise, achieving accurate label noise separation.}
	\label{fig_histmao}
\end{figure}

In previous works~\cite{kulhanek2024wildgaussians,sabour2024spotlesssplats,wang2024we,chen2024nerf}, anti-noise reconstruction was commonly framed as a metric-classification problem. These methods use reconstructed residuals or uncertainty values to quantify noise and design a metric-based classifier to separate noisy labels from clean data. However, in different scenarios, the types and proportions of noise labels, as well as the limitations of the observation itself, can affect the distribution of noise and clean background. The previous naive metric classification strategy is insufficient to handle real-world scene variability. To address the above issues, we propose the techniques of \textbf{observation cognitive correction} and \textbf{dynamic anchor point combination thresholding}.

\textit{Observation Cognitive Correction.}
Consider the question: \textbf{Are the captured images truly sufficient to reconstruct every region of the scene perfectly?} In classic datasets~\cite{ren2024nerf,barron2022mip,sabour2023robustnerf}, areas outside the primary view are often far away and observed fewer times. Even without occlusion, reconstruction errors are still significant in these regions. As shown in Fig.~\ref{fig_histmao}, both the red and blue boxed areas are clean, but the blue box, with fewer observations, exhibits more significant reconstruction errors than the red box. This indicates that sufficient observation is a crucial factor for both reconstruction and noise label separation.

Inspired by the principle of triangulation~\cite{hartley1997triangulation,thurmond2001point}, we quantified the camera's observation of Gaussian primitives in 3DGS reconstruction, called observation completeness (OC), which is proportional to the number of effective observations and the diversity of observation camera angles. Furthermore, we propose a noise assessment correction scheme based on OC, which balances measurement errors caused by insufficient observations and significantly improves the generalization of the noise classifier.

\textit{Dynamic Anchor Combination Threshold.} The proportion of noise and the difficulty of scene reconstruction significantly affect the statistical distribution of background and noisy foreground. Therefore, we propose a dynamic threshold anchor scheme. The noise foreground and background classification threshold for each scene is calculated by two statistical anchor points. They are the maximum inter-class variance position~\cite{otsu1975threshold} and the mean position of the background class. Unlike the fixed threshold methods~\cite{kulhanek2024wildgaussians,sabour2024spotlesssplats}, our threshold is calculated based on the unique statistical characteristics of each scene, so it can be directly applied to most scenes without adjusting any parameters. In Tab.~\ref{tab:ablation}, Tab.~\ref{tab:onthego} and Fig.~\ref{statue}, we proved that our dynamic threshold scheme could achieve much better noise separation and scene reconstruction effects even though scene characteristics are vastly different.

Finally, we developed our method OCSplats based on the 3DGS~\cite{kerbl20233d}.
The overall pipeline can be summarized as reconstruction noise assessment, observation completeness-based correction, and self-supervised label noise classification based on dynamic thresholds. 
Moreover, we propose a pruning strategy for 3DGS based on observation completeness. The core idea is to remove Gaussian primitives with severely insufficient effective observations. This strategy not only reduces the total number of Gaussian primitives by approximately 10\%, but also significantly stabilizes the quality of 3DGS reconstruction.
Through experiments on mainstream datasets with noisy reconstructions, OCSplats achieves the best visual performance in label noise classification and demonstrates significantly superior reconstruction accuracy, especially in complex scenes.

Our key contributions can be summarized as follows:
\begin{enumerate}
	\item A novel quantitative measure of observation completeness (OC) method, which allows people to visually determine whether the observation of each region in the scene is sufficient.
	\item An assessment correction and pruning method based on observation completeness was proposed, significantly improving the erroneous noise classification caused by lack of observation.
	\item A noise classifier self-supervised training pipeline based on dynamic threshold anchor points, achieving robust noise label classification in scenes with different complexities and occlusions.
	
\end{enumerate}

%% file: sec/2_relateworks.tex
\section{Related Works}

\begin{figure*}[!t]
	\centering
	\includegraphics[width=6.8in]{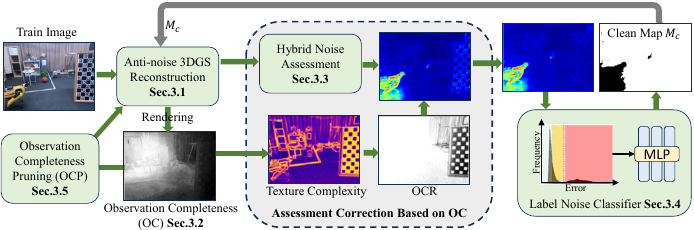}
	\caption{\textbf{Pipeline} Our method is mainly divided into two stages. In the first stage, we assess the noise present in the labels. In the second stage, we self-supervised train an MLP network to infer reasonable clean regions from feature embeddings. Specifically, this paper proposes a learning completeness theory for 3DGS to modify noise assessment, which can more accurately evaluate and identify noise in different scenarios.}
	\label{fig_method}
\end{figure*}

\subsection{Epistemic Uncertainty and Reconstruction} 
Generally, epistemic uncertainty~\cite{swiler2009epistemic,hullermeier2021aleatoric,jakeman2010numerical,nguyen2019epistemic} mainly arises from insufficient training data. For example, in detection tasks, the detector cannot recognize objects it has never encountered before. In the context of scene reconstruction, epistemic uncertainty typically originates from the limited shooting perspective~\cite{barbano2020quantifying,narnhofer2021bayesian}. Shen et al.~\cite{shen2024estimating} quantified epistemic uncertainty using ensemble learning to visualize unobserved regions in NeRF reconstructions. Similarly, Goli et al.~\cite{Goli_2024_CVPR} quantifies the location uncertainty of NeRF voxels by actively introducing perturbations and constructing an uncertainty field.

In this paper, observation completeness (OC) is similar to epistemic uncertainty, but our OC focuses more on whether the scene appearance has been adequately learned. In contrast to prior work~\cite{shen2024estimating,Goli_2024_CVPR}, our OC employs an iterative computation method tailored for 3DGS, making it highly efficient without requiring extensive post-processing. Additionally, we apply OC to noise-robust reconstruction, effectively helping 3DGS improve reconstruction quality and reduce label noise classification errors.

\subsection{Stochastic Uncertainty and Reconstruction} 
Stochastic uncertainty~\cite{maddox2019simple} arises from the inherent noise in the data itself and does not diminish even as the data size grows. Most existing NeRF-based denoising reconstruction methods utilize the stochastic uncertainty loss~\cite{kendall2017uncertainties,ren2024nerf} to supervise the NeRF reconstruction. In this setting, the stochastic uncertainty loss actively suppresses errors in regions that are difficult to learn—such as moving objects, non-Lambertian surfaces, or areas lacking sufficient observations—thereby achieving a noise-robust learning effect.

However, in the 3DGS method, Gaussian split pruning relies on gradients. The stochastic uncertainty loss significantly shifts the loss weighting, which ultimately causes 3DGS to collapse. Consequently, existing approaches~\cite{kulhanek2024wildgaussians} block the back-propagation of the uncertainty loss and adopt a metric-threshold classification strategy to identify noisy labels. A similar residual-based method~\cite{sabour2024spotlesssplats} also exists, differing only in that it uses reconstruction residuals rather than stochastic uncertainty as its threshold classification metric. The core problem with these threshold-based methods is their inability to accommodate diverse scenarios: each scenario requires retuning the classification threshold to achieve ideal performance. Our proposed dynamic anchor threshold approach circumvents threshold-based methods' poor generalizability while maintaining high reconstruction quality.

%% file: sec/3_method.tex
\section{Our Approach}
This section introduces how to implement our method, and the overall process is shown in Fig.~\ref{fig_method}. Specifically, Sec.\ref{MA} introduces key background knowledge about 3DGS reconstruction; Sec.\ref{MB} proposes quantification and visualization methods for observation completeness; Sec.\ref{MC} proposes a hybrid noise assessment scheme and how to perform assessment correction based on observation completeness; Sec.\ref{MD} proposes a self-supervised separation scheme for label noise based on hybrid assessment; Sec.\ref{MF} introduces a novel pruning strategy based on observation completeness.

\subsection{Background Knowledge of 3DGS} \label{MA}

Our method is implemented by 3DGS technology, where the scene is represented as a set of explicit 3DGS primitives $\mathcal{G}= \{g_1,g_2,...,g_N \}$. Each Gaussian primitive $g$ is defined by its center position $\mathbf{p}_k$, pose parameter $\boldsymbol{\Sigma}_k$, opacity parameter $\alpha_k \in[0,1]$, and spherical harmonic parameter $\mathbf{SH}$ with respect to color $\mathbf{c}$. During rendering, the 3D Gaussian primitive is first projected onto the imaging plane, and the projected 2D Gaussian primitive $g_k^{2D}$ can be written as:
\begin{equation}
	g_k^{2D}(\bm{x})=e^{-\frac{1}{2}\left(\bm{x}-\mathbf{p}_k\right)^T  \left(\boldsymbol{\Sigma}_k^{2 D}\right)  ^{-1}\left(\bm{x}-\mathbf{p}_k\right)}
\end{equation}
where $\boldsymbol{\Sigma}_k^{2 D}$ is a 2D covariance matrix calculated by $\boldsymbol{\Sigma}_k$.

Next, 3DGS calculates the color $\bm{\mathbf{C}(x)}$ of pixels via alpha blending according to the primitive’s depth order.
\begin{equation}
	\bm{\mathbf{C}(x)}=\sum_{k=1}^K w_k\mathbf{c}_k 
	\label{eq2}
\end{equation}
\begin{equation}
	w_k= \alpha_k g_k^{2D}(\bm{x}) \prod_{j=1}^{k-1}\left(1-\alpha_j g_j^{2D}(\bm{x})\right)
\end{equation}
among them, $\mathbf{c}_k $ is a view-dependent color decoded from spherical harmonic parameter~\cite{seeley1966spherical} $\mathbf{SH}$. 

Finally, 3DGS optimizes the parameters contained in the Gaussian primitive through photometric and SSIM mixed loss:
\begin{equation}
	\mathcal{L}_{gs}=\lambda_{\mathrm{s}} \operatorname{SSIM}(\bm{\mathbf{C}}, \bm{\mathbf{C}_{gt}})+\left(1-\lambda_{\mathrm{s}}\right)\|\bm{\mathbf{C}}-\bm{\mathbf{C}_{gt}}\|_1
\end{equation}
where $\bm{\mathbf{C}}$ is the rendered color, $\bm{\mathbf{C}_{gt}}$ is the ground-truth color, and $\operatorname{SSIM}$~\cite{wang2004image} is the structural similarity function.

\textbf{Anti-noise Reconstruction.}
In order to prevent the impact of noise labels on reconstruction, the usual anti-noise reconstruction method~\cite{kulhanek2024wildgaussians,sabour2024spotlesssplats} estimates a clean mask $M_c$ to filter the reconstruction loss:
\begin{equation}
	\mathcal{L}_{final}=M_c \mathcal{L}_{gs}
\end{equation}
In this paper, we also use $M_c$ to achieve anti-noise reconstruction. Compared to the uncertainty loss~\cite{kendall2017uncertainties,ren2024nerf} used in NeRF, a masking scheme with accurate noise segmentation can achieve more accurate reconstruction results.

\subsection{Observation  Completeness} \label{MB}
This section proposes a calculation and rendering method for observation completeness (OC), which is an indicator constructed based on the principle of triangulation. It describes whether Gaussian primitives have obtained sufficient observations to ensure correct reconstruction.
Specifically, the observation completeness metric $O_i^n$ of the Gaussian primitive $g_i$ at the $n$th iteration is defined as follows:

\begin{equation}
	O_i^n=\lambda_{oc} O_i^{n-1} + (1-\lambda_{oc})\delta_i^n
\end{equation}
\begin{equation}
	\delta_i={\hat{\sigma}^i(n) u_i^n}
	\label{eq66}
\end{equation}
\begin{equation}
	u_i^n= \begin{cases}1 & Gradp(g_i^n)>1\times 10^{-7} \\ 0 & \text { otherwise }\end{cases}
\end{equation}
where $\lambda_{oc}=0.98$, $\hat{\sigma}^i(n)$ represents the positional variance between cameras observing the Gaussian primitive $g_i$, which reveals whether the observation perspective of $g_i$ is sufficient.
$u_i^n$ determines whether the Gaussian primitive $g_i$ is effectively observed in the current training frame, and we set $u_i^n$ to 1 when the positional gradient $Gradp(g_i^n)$ is greater enough.

In Eq.\ref{eq66}, $u_i^n$ can be directly obtained during a single training iteration, but $\hat{\sigma}^i(n)$ requires computing the variance of historical camera positions, which significantly increases computational resource consumption. Therefore, we use an iterative formula for variance to compute $\hat{\sigma}^i(n)$:
\begin{equation}
	\hat{\mu}_{m}^i=\hat{\mu}_{m-1}^i+\frac{1}{m}\left(T_{n}-\hat{\mu}_{m-1}^i\right)
	\label{xita1}
\end{equation}
\begin{equation}
	\hat{\sigma}_{m}^i=\left(\frac{m-2}{m-1}\right) \hat{\sigma}_{m-1}^i+\frac{1}{m}\left(T_{n}-\hat{\mu}_{m-1}^i\right)^2
	\label{xita2}
\end{equation}
\begin{equation}
	m = m + u_i^n
	\label{xita3}
\end{equation}
where $\hat{\sigma}^i(n) = \|\hat{\sigma}_{m}^i\|_2 $, Eq.\ref{xita1}, Eq.\ref{xita2} and Eq.\ref{xita3} are executed if and only if $u_i^n=1$, so $m$ is not greater than $n$. $T_{n}$ is the spatial position of the camera in the $n$th iteration.

\textbf{Rendering of OC.} Referring to the color rendering  in Eq.~\ref{eq2}, the rendering for OC is as follows:
\begin{equation}
	\bm{\mathbf{O}(x)}=\sum_{k=1}^K w_k O_k 
	\label{eqoc}
\end{equation}


\subsection{Hybrid Noise Assessment and Correction} \label{MC}

\begin{figure}[!t]
	\centering
	\includegraphics[width=3.2in]{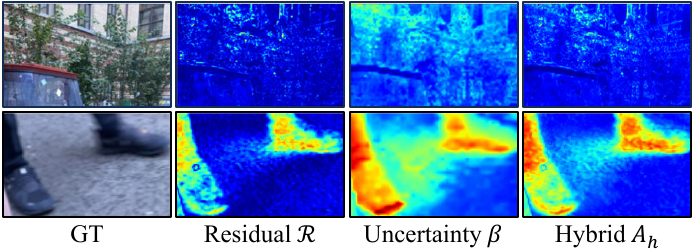}
	\caption{\textbf{Hybrid Noise Assessment.} Hybrid assessment can reduce additional noise caused by reconstruction difficulties (up) while focusing on persistent little errors (down).}
	\label{fig_mix}
\end{figure}
\begin{figure}[!t]
	\centering
	\includegraphics[width=3.2in]{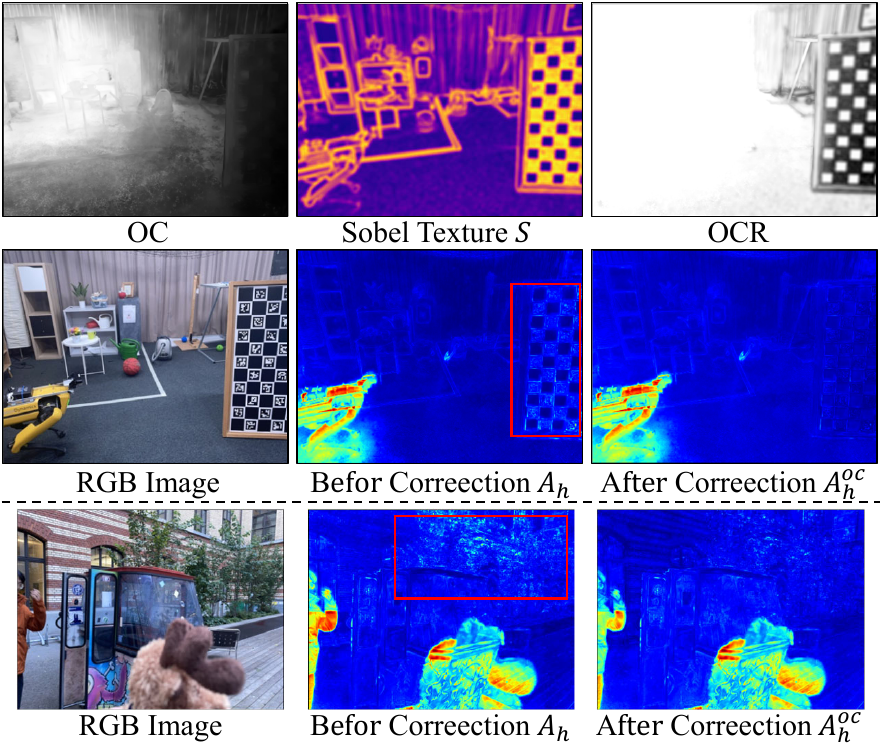}
	\caption{\textbf{Observation Correction.} The observation correction ratio $OCR$ is determined by considering the completeness of observations $\mathbf{O}$ in a certain region and its reconstruction difficulty $S$.}
	\label{fig_ocr}
\end{figure}

We propose a hybrid noise evaluation scheme consisting of residual $\mathcal{R}$ and uncertainty $\beta$. As shown in Fig.~\ref{fig_mix}, $\mathcal{R}$ reflects the significant noise present in the current training stage, while $\beta$ reflects the persistent interference. However, $\mathcal{R}$ is difficult to detect weaker noise (foot shadows), and $\beta$ cannot distinguish difficult-to-reconstruct areas (distant trees), all of which interfere with accurate noise separation. Our proposed simple and effective hybrid assessment $\mathcal{A}_{h}$ can avoid these drawbacks:
\begin{equation}
	\mathcal{A}_{h}=\lambda_2 \mathcal{R} +(1-\lambda_2) \beta
\end{equation}
where $\lambda_2=0.5$, $\mathcal{R}=\|\bm{\mathbf{C}}-\bm{\mathbf{C}_{gt}}\|_1$. As shown in Fig.~\ref{fig_mix}, the hybrid assessment enhances the error region shared by the residuals $\mathcal{R}$ and $\beta$, indirectly suppressing the interference caused by learning difficulties in $\beta$. We elaborated on the calculation process of $\beta$ in detail in Sec \ref{uncer}.

\textbf{Observation Cognitive Correction.}
The example in Fig.~\ref{fig_histmao} shows that in 3DGS reconstruction, there is a high correlation between the number of observations and the reconstruction error. Therefore, we further correct the label noise assessment using the observation completeness proposed in Eq.\ref{eqoc}:
\begin{equation}
	\mathcal{A}_{h}^{oc}= \mathcal{A}_{h} \cdot OCR
\end{equation}
\begin{equation}
	OCR= 1-\lambda_3(th-\mathbf{O})S
\end{equation}
where $OCR$ represents the observation correction ratio, with its range constrained between 0 and 1. $\lambda_3$ and $th$ are hyperparameters set to 3.0 and 0.3, respectively. $\mathbf{O}$ denotes the observation completeness, whose range is constrained between 0 and $th$. $S$ represents the texture intensity, calculated by convolving the image with the Sobel~\cite{gao2010improved} operator.

The process of correcting the observation is shown in the upper part of Fig.~\ref{fig_ocr}. After correcting based on observation completeness $\mathbf{O}$, the errors caused by the lack of observation (red box in Fig.~\ref{fig_ocr}) in $\mathcal{A}_{h}$ are significantly suppressed. This enables subsequent classifiers to learn the correct static scene mask more robustly from $\mathcal{A}_{h}^{oc}$.

\begin{figure}[!t]
	\centering
	\includegraphics[width=3.2in]{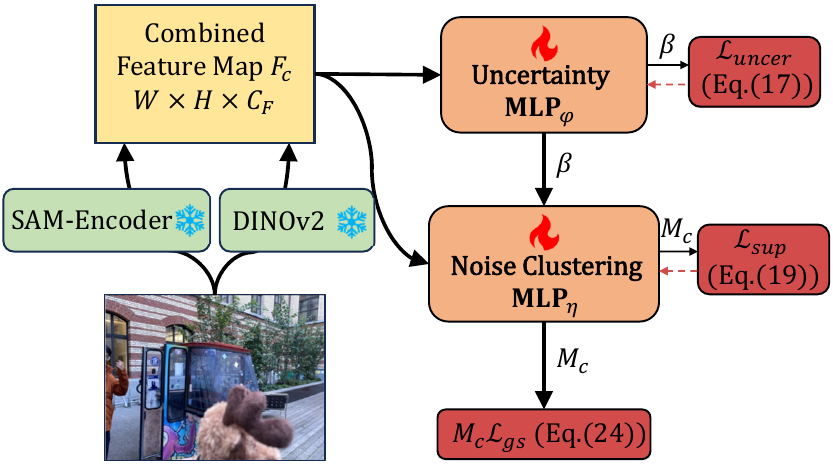}
	\caption{\textbf{Label Noise classification Pipeline.} The pre-trained feature extractor (DINOv2,SAM) extracts feature maps from the training images. Then, the uncertainty $\mathbf{MLP}_{\psi}$ and noise classifier $\mathbf{MLP}_{\eta}$ predict the uncertainty $\beta$ and static background mask $M_c$ from the feature map. Three losses (right side) are used to optimize $\mathbf{MLP}_{\psi}$, $\mathbf{MLP}_{\eta}$, and 3DGS model.}
	\label{fig_class}
\end{figure}
\begin{figure}[!t]
	\centering
	\includegraphics[width=3in]{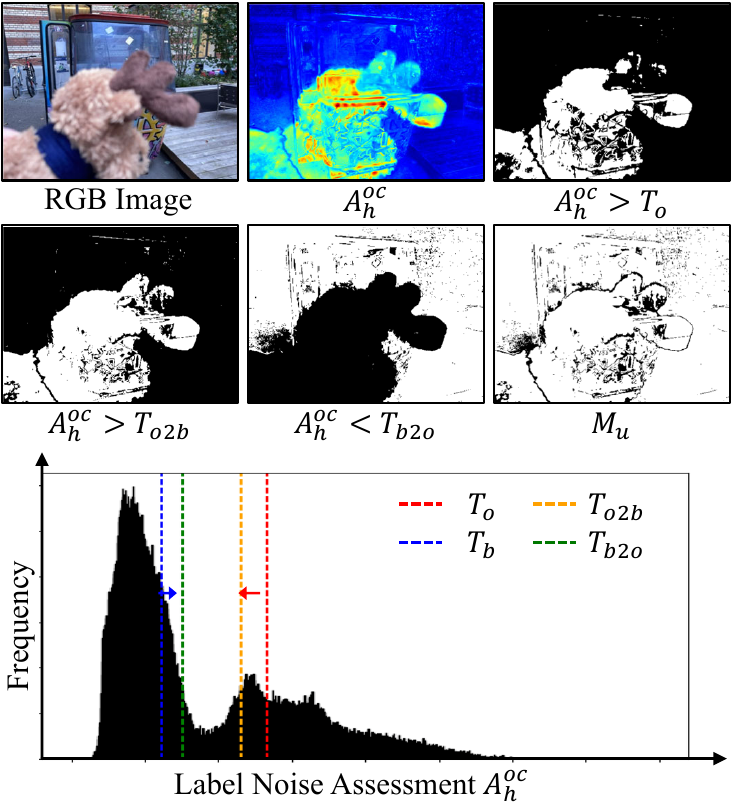}
	\caption{\textbf{Dynamic Threshold Clustering.} Using two dynamic anchors, the maximum inter-class variance (red) and the background noise mean (blue), we can stably calculate the threshold intervals for foreground and background.}
	\label{Aclass}
\end{figure}
\subsection{Label Noise Classifier} \label{MD}
As shown in Fig.~\ref{fig_class}, we train two MLPs to learn clean mask $M_c$ and uncertainty $\beta$ from image feature encoding. One significant advantage of using an MLP for learning is that predicting masks from features with clear semantics enables a better understanding of chunked semantic regions, further avoiding fragmented noise.

\textbf{Feature Extraction.} 
We use SAM\cite{kirillov2023segment} and DINOv2\cite{oquab2023dinov2} with frozen weights to extract features from the training images. Then, we scale the features to the original image size and concatenate them together to obtain $F_c$. Moreover, we also concatenated position encoding $F_{pos}$ on $F_c$ to assist MLP in learning uncertainty in low feature response areas. 

\begin{table*}[htbp]
	\centering
	\caption{\textbf{Ablation Study.} We conducted ablation in different scenes, including indoor scenes with moderate (\textit{Statue}) and high (\textit{Spot}) occlusion, and outdoor scene (\textit{Fountain}) with less occlusion but containing reflections.}
	\begin{tabular}{llcccccccc}
		\toprule
		&       & \multicolumn{2}{c}{\textit{Statue}} & \multicolumn{2}{c}{\textit{Fountain}} & \multicolumn{2}{c}{\textit{Spot}} & \multicolumn{2}{c}{\textit{Mean}} \\
		& ID    & PSNR$\uparrow$  & SSIM$\uparrow$  & PSNR$\uparrow$  & SSIM$\uparrow$  & PSNR$\uparrow$  & SSIM$\uparrow$  & PSNR$\uparrow$  & SSIM$\uparrow$ \\
		\midrule
		\textbf{Ours} & \textbf{A}     & 23.13 & 0.863 & \textbf{23.05} & \textbf{0.794} & \textbf{26.32} & \textbf{0.902} & \textbf{24.17} & \textbf{0.853} \\
		only uncertainty $\beta$  & \textbf{B}     & 22.85 & 0.862 & 19.97 & 0.691 & 26.31 & 0.901 & 23.04 & 0.818 \\
		only residuals $\mathcal{R}$ & \textbf{C}     & \textbf{23.16} & \textbf{0.867} & 23.01 & 0.782 & 25.21 & 0.899 & 23.79 & 0.849  \\
		w/o OCP & \textbf{D}     & 23.14  & 0.863  & 22.78 & 0.792  & 26.24  & 0.901  & 24.05 & 0.852  \\
		w/o OCC & \textbf{E}     & 23.03 & 0.865 & 22.85 & 0.794 & 25.72 & 0.901 & 23.86 & 0.853  \\
		fixed threshold & \textbf{F}     & 22.90 & 0.862 & 22.87 & 0.791 & 25.60 & 0.899 & 23.78 & 0.850  \\
		\bottomrule
	\end{tabular}%
	\label{tab:ablation}%
\end{table*}%

\textbf{Learning Uncertainty.} \label{uncer}
Unlike the Nerf methods~\cite{ren2024nerf,martin2021nerf}, introducing uncertainty $\beta$ will greatly affect the splitting and pruning of 3D Gaussian primitive, leading to training collapse. Therefore, we blocked the gradient feedback related to Gaussian primitive parameters, and the final uncertainty loss $\mathcal{L}_{uncer}$ is as follows:
\begin{equation}
	\beta=\mathbf{MLP}_{\psi}(F_{pos},F_{id},F_c)
	\label{mlpuncer}
\end{equation}
\begin{equation}
	\mathcal{L}_{uncer}=\frac{ Detach(\mathcal{L}_{gs})}{2 \beta^2}+ \log \beta
	\label{equncer}
\end{equation}
where $Detach$ is a gradient blocker (Similar to $.detach()$ in pytorch~\cite{paszke2019pytorch}), $F_{id}$ is the image number encoding, $\psi$ is the network parameter to be learned.

\textbf{Learning Noise Classifier.} We train a noise classifier $\mathbf{MLP}_{\eta}$ by generating self-supervised labels from $\mathcal{A}_{h}^{oc}$:
\begin{equation}
	M_c=\mathbf{MLP}_{\eta}(F_{pos},\beta,F_c)
	\label{mlpclean2}
\end{equation}
\begin{equation}
	\mathcal{L}_{sup}= M_{u} \| M_c - M_{self}\|_1
	\label{eqclean1}
\end{equation}
among them, $M_{self}$ and $M_{u}$ are the self-supervised labels and effective training regions, respectively. Next, we introduce how to calculate the labels required for self-supervision from $\mathcal{A}_{h}^{oc}$.

\textbf{Self-supervised Labels.}
We propose an dynamic threshold anchor method to handle various scenarios. Specifically, we first convert $\mathcal{A}_{h}^{oc}$ into a histogram to analyze the occurrence frequencies of noise with different magnitudes statistically. Then, we compute the maximum inter-class variance threshold $T_o$ and the mean of the background noise $T_b$ as dynamic classification anchors. Finally, the self-supervised labels required for training are calculated using a weighted combination of these dynamic anchors:
\begin{equation}
	T_{b2o}=  (1-\lambda_4) T_{b} + \lambda_4 T_o
	\label{o2b3}
\end{equation}
\begin{equation}
	T_{o2b}=  \lambda_5 T_{b} + (1-\lambda_5)T_o
	\label{o2b4}
\end{equation}
\begin{equation}
	M_{self}=  \mathcal{A}_{h}^{oc}<T_{b2o}
	\label{o2b}
\end{equation}
\begin{equation}
	M_{u}=  (\mathcal{A}_{h}^{oc}<T_{b2o}) \lor (\mathcal{A}_{h}^{oc}>T_{o2b})
	\label{o2bu}
\end{equation}
where $T_{b2o}$ and $T_{o2b}$ are the fine-tuned foreground and background thresholds, $M_{self}$ is the generated self-supervised label, and $M_u$ is the effective training area. Please refer to the \textbf{supplementary materials} for the specific calculation process of $T_b$ and $T_o$.

Fig.~\ref{Aclass} shows the complete process of label generation. By leveraging $M_u$, the MLP focuses on learning the more prominent errors while minimizing the interference caused by ambiguous intermediate regions during training. Moreover, the maximum inter-class variance threshold demonstrates stable performance in the noise classification task proposed in this paper. Through experiments in Tab.~\ref{tab:ablation} and Tab.~\ref{tab:onthego}, we show that, compared to fixed-threshold classification, our dynamic thresholding approach can seamlessly adapt to most scenarios without parameter adjustments.

\textbf{Parameter Optimization.}
Finally, we optimize our OCSplats through the following loss:
\begin{equation}
	\mathcal{L}_{all}=M_c \mathcal{L}_{gs} + \mathcal{L}_{uncer} + \mathcal{L}_{sup}
	\label{eqall}
\end{equation}

\begin{figure}[!t]
	\centering
	\includegraphics[width=2.7in]{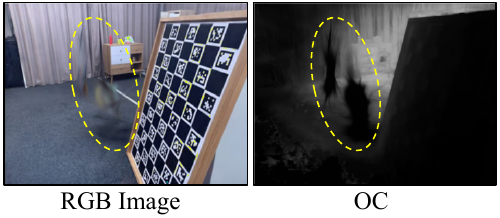}
	\caption{\textbf{Observation Completeness Pruning (OCP).} The OC values of floating objects are often very low, making it easy to distinguish them from normal pixels.}
	\label{cgr}
\end{figure}

\subsection{Observation Completeness Pruning (OCP)} \label{MF}
After visualizing the observation completeness, we also found an interesting phenomenon. As shown in Fig.~\ref{cgr}, the observation completeness of floating Gaussians (floaters) was extremely low. It is easy to understand. The fundamental reason why the floating Gaussian is not optimized during training is that it is only represented in a small number of viewing angles and invisible in the vast majority of viewing angles.
Based on the above observations, we propose a pruning strategy based on observation completeness. Specifically, during the training process, we directly eliminate those Gaussians whose observation completeness is less than 0.03 ($O_i^n<0.03$) and whose number of observations is less than 3 (during a full round of training). 


%% file: sec/4_exp.tex
\section{Experiment}

\begin{table*}[htbp]
	\centering
	\caption{\textbf{Evaluation on RobustNeRF Dataset.} We presented a quantitative comparison between our method and the baseline method, with the upper half being based on the NeRF method and the lower half being based on the 3DGS method. The best method between classes is displayed in bold.}
	\setlength{\tabcolsep}{2.2pt}
	\begin{tabular}{lcccccccccccc}
		\toprule
		& \multicolumn{3}{c}{\textit{Android}} & \multicolumn{3}{c}{\textit{Crab}} & \multicolumn{3}{c}{\textit{Yoda}} & \multicolumn{3}{c}{\textit{Statue}} \\
		& PSNR$\uparrow$  & SSIM$\uparrow$  & LPIPS$\downarrow$ & PSNR$\uparrow$  & SSIM$\uparrow$  & LPIPS$\downarrow$ & PSNR$\uparrow$  & SSIM$\uparrow$  & LPIPS$\downarrow$ & PSNR$\uparrow$  & SSIM$\uparrow$  & LPIPS$\downarrow$ \\
		\midrule
		Mip-NeRF360~\cite{barron2022mip} & 21.81 & 0.695  & 0.176  & 29.25 & 0.918  & 0.086  & 23.75 & 0.770  & 0.216  & 19.86 & 0.690  & 0.233  \\
		NeRF-W~\cite{martin2021nerf} & 20.62  & 0.664  & 0.258  & 26.91  & 0.866  & 0.157  & 28.64  & 0.752  & 0.260  & 18.91  & 0.616  & 0.369  \\
		RobustNeRF~\cite{sabour2023robustnerf} & 23.28  & 0.755  & 0.130  & 32.22  & 0.945  & \textbf{0.060} & 29.78  & 0.821  & 0.150  & 20.60  & 0.758  & 0.150  \\
		NeRF-HuGS~\cite{chen2024nerf} & 23.32  & 0.763  & 0.200  & 34.16  & 0.956  & 0.070  & 30.70  & 0.834  & 0.220  & 21.00  & 0.774  & 0.180  \\
		\midrule
		SpotLessSplats~\cite{sabour2024spotlesssplats} & 24.83  & 0.826  & 0.085  & 34.43 & 0.953  & 0.079  & 35.11  & 0.956  & 0.073  & 22.24  & 0.828  & 0.136  \\
		3DGS~\cite{kerbl20233d}  & 22.65 & 0.810  & 0.130  & 31.13  & 0.936  & 0.102  & 26.94 & 0.910  & 0.140  & 21.02 & 0.810  & 0.160  \\
		OCSplats (\textbf{Ours})  & \textbf{25.17} & \textbf{0.843} & \textbf{0.071} & \textbf{35.21} & \textbf{0.958} & 0.068 & \textbf{35.37} & \textbf{0.961} & \textbf{0.069} & \textbf{23.13} & \textbf{0.863} & \textbf{0.101} \\
		\bottomrule
	\end{tabular}%
	\label{tab:robust}%
\end{table*}%

In this section, we first conduct ablation experiments on the key modules proposed in this paper to verify their effectiveness. Next, we compare OCSplats with SOTA methods on interference datasets. 

\textbf{Datasets.} 
We mainly evaluated our method on two datasets, RobustNeRF~\cite{sabour2023robustnerf} and On-the-go~\cite{ren2024nerf}. The RobustNeRF consists of four scenes of toys-on-the-tables, with low occlusion rates for all scenes except the \textit{Statue}. The On-the-go dataset consists of outdoor and indoor scenes. Referring to previous work~\cite{ren2024nerf}, we selected 6 sequences in descending order of occlusion rate. In addition, compared to the RobustNeRF dataset, the On-the-go dataset typically has lower observational completeness.

\textbf{Implementation Details.} 
We stacked the key modules proposed in this paper on the standard 3DGS method. The difference is that our OCSplats run OCP and Gaussian pruning clone every $z$ steps ($z$ is the length of the scene ).

\begin{figure}[!t]
	\centering
	\includegraphics[width=3.2in]{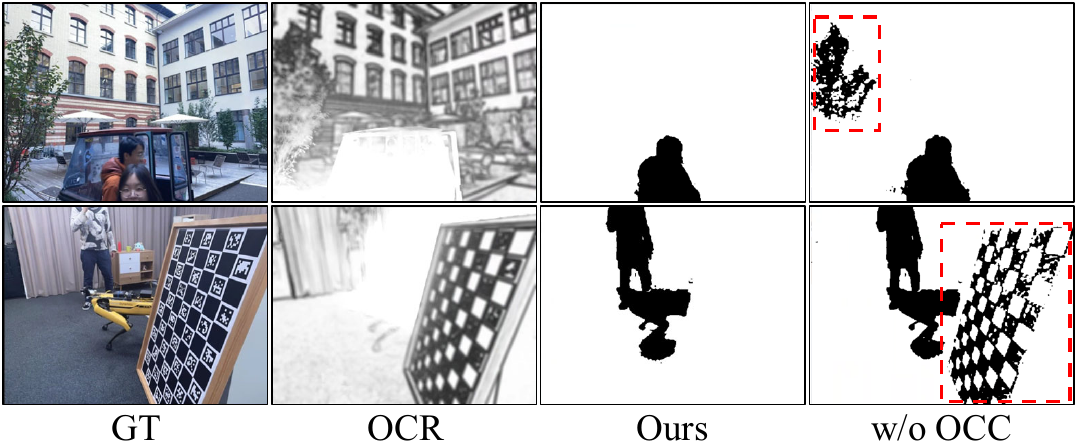}
	\caption{\textbf{Ablation Study about OCC.} We visualized and ablated the observation completeness correction (OCC) module. After using OCC, the noise classification performance in areas with fewer observations was significantly improved. }
	\label{ablation}
\end{figure}
\subsection{Ablation Study}

\textbf{Hybrid Assessment.} 
To validate the effectiveness of the hybrid assessment, we separately tested the performance of residual $\mathcal{R}$ (\textbf{C}) and uncertainty $\beta$ (\textbf{B}). As shown in Tab.~\ref{tab:ablation}, when only residual $\mathcal{R}$ was used, the model performed slightly worse in the shadow-heavy \textit{Spot} scenario. In parallel, when only uncertainty $\beta$ was applied, the model experienced a performance collapse in the \textit{Fountain} (have many non-Lambertian surfaces) scene (23.05 to 19.97).
On the contrary, our hybrid assessment (\textbf{A}) achieved the optimal average performance, which proves that our hybrid strategy can preserve the advantages of each indicator as much as possible while reducing the impact of its defects on overall performance.

\begin{table*}[htbp]
	\centering
	\caption{\textbf{Evaluation on On-the-go Dataset.} The \textbf{SpotLessSplasts} in the table are the results we reproduced; We used default fixed thresholds in all scenarios.}
	\setlength{\tabcolsep}{2.4pt}
	\begin{tabular}{lcccccccccccc}
		\toprule
		& \multicolumn{4}{c}{Low Occlusion} & \multicolumn{4}{c}{Medium Occlusion} & \multicolumn{4}{c}{High Occlusion} \\
		& \multicolumn{2}{c}{\textit{Mountain}} & \multicolumn{2}{c}{\textit{Fountain}} & \multicolumn{2}{c}{\textit{Corner}} & \multicolumn{2}{c}{\textit{Patio}} & \multicolumn{2}{c}{\textit{Spot}} & \multicolumn{2}{c}{\textit{Patio high}} \\
		& PSNR$\uparrow$  & SSIM$\uparrow$  & PSNR$\uparrow$  & SSIM$\uparrow$  & PSNR$\uparrow$  & SSIM$\uparrow$  & PSNR$\uparrow$  & SSIM$\uparrow$  & PSNR$\uparrow$  & SSIM$\uparrow$  & PSNR$\uparrow$  & SSIM$\uparrow$ \\
		\midrule
		Mip-NeRF360~\cite{barron2022mip} & 19.64  & 0.601  & 13.91 & 0.290  & 20.41  & 0.660  & 15.48  & 0.503  & 17.82  & 0.306  & 15.73  & 0.432  \\
		NeRF-W~\cite{martin2021nerf} & 18.07  & 0.492  & 17.20  & 0.410  & 20.21  & 0.708  & 17.55  & 0.532  & 16.40  & 0.384  & 12.99  & 0.349  \\
		NeRFOn-the-go~\cite{ren2024nerf} & 20.15  & 0.644  & 20.11  & 0.609  & 24.22  & 0.806  & 20.78  & 0.754  & 23.33  & 0.787  & 21.41  & 0.718  \\
		RobustNeRF~\cite{sabour2023robustnerf} & 17.54  & 0.496  & 15.65  & 0.318  & 23.04  & 0.764  & 20.39  & 0.718  & 20.65  & 0.625  & 20.54  & 0.578  \\
		\midrule
		Wild-GS~\cite{kulhanek2024wildgaussians} & 20.92  & 0.673  & 20.94 & 0.672  & 23.69  & 0.815  & 21.23  & 0.805  & 23.93  & 0.781  & 22.11  & 0.737  \\
		3DGS~\cite{kerbl20233d}  & 20.98  & 0.709  & 21.46 & 0.728  & 22.54  & 0.800  & 17.13  & 0.671  & 18.24  & 0.657  & 15.74  & 0.515  \\
		SpotLessSplats~\cite{sabour2024spotlesssplats} & 21.39  & 0.708  & 22.62  & 0.761  & 26.09  & 0.877  & 22.28  & 0.823  & 24.15  & 0.819  & 22.27  & 0.771  \\
		OCSplats(\textbf{Ours})  & \textbf{21.45} & \textbf{0.738} & \textbf{23.05} & \textbf{0.794} & \textbf{26.74} & \textbf{0.904} & \textbf{22.45} & \textbf{0.841} & \textbf{26.32} & \textbf{0.902} & \textbf{23.61} & \textbf{0.846} \\
		\bottomrule
	\end{tabular}%
	\label{tab:onthego}%
\end{table*}%

\textbf{Observation Completeness Pruning (OCP).}
Unlike SplotLessSplats~\cite{sabour2024spotlesssplats} that crudely removed low-gradient Gaussian elements (which resulted in performance degradation), OCP removes Gaussian primitives that cannot be reconstructed correctly from the triangulation perspective, which can further improve overall performance while pruning. As shown in Tab.~\ref{tab:ablation} (\textbf{D}), the average performance decreased from 24.17 to 24.05 after removing OCP pruning.

\begin{figure}[!t]
	\centering
	\includegraphics[width=3.2in]{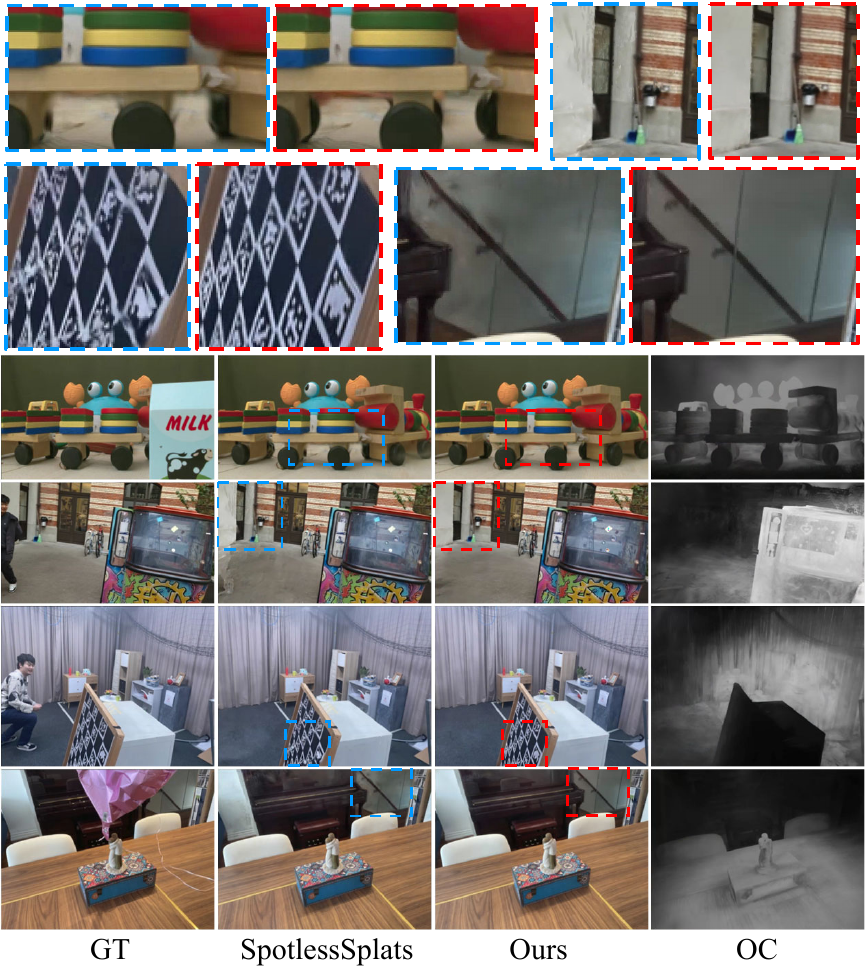}
	\caption{\textbf{Visual Comparison of Rendering Results.} Compared with the state-of-the-art method (Blue box), our approach (Red box) achieved better rendering results in areas lacking observation.}
	\label{rgbshow}
\end{figure}

\textbf{Observation Completeness Correction (OCC).}
As shown in Tab.~\ref{tab:ablation} (\textbf{E}), the average performance significantly  decreased after further ablation of OCC. Moreover, OCC is of great significance for separating label noise. As shown in Fig.~\ref{ablation} and Fig.~\ref{rgbshow}, using OCC can significantly improve the noise separation effect and rendering performance in areas with missing observations.

\textbf{Dynamic Threshold.}\label{dt}
To verify the effectiveness of the dynamic threshold in Sec.\ref{MD}, we attempted a fixed threshold version based on Goli et al.'s work~\cite{sabour2024spotlesssplats}. Specifically, we consider the pixels with the last 50\% error as static background and those with the first 10\% as noise foreground.
As shown in Tab.~\ref{tab:ablation} (\textbf{F}), the average PSNR decreased from 24.17 to 23.78, and the performance degradation was more pronounced in high-occlusion scenes (\textit{Spot}). This demonstrates the importance of our dynamic thresholding strategy for anti-noise reconstruction in complex scenes.

\begin{figure}[!t]
	\centering
	\includegraphics[width=3.2in]{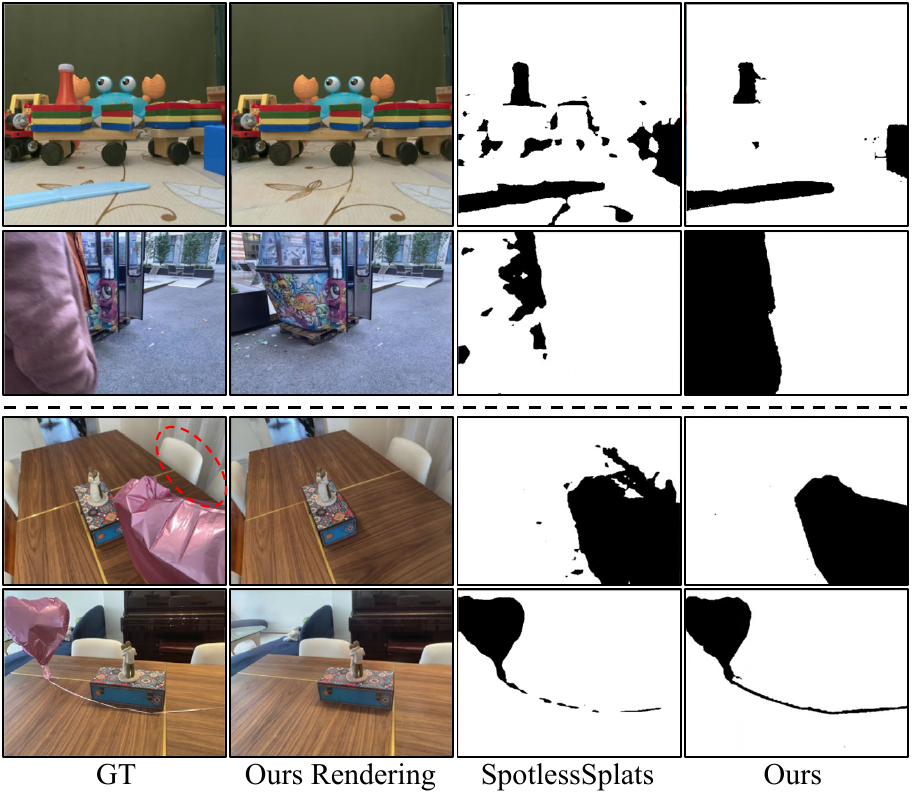}
	\caption{\textbf{Noise Label Segmentation.} Compared to the previous method of using a fixed threshold, our approach can robustly segment the correct static background in various complex scenarios.}
	\label{statue}
\end{figure}


\subsection{Evaluation on RobustNeRF Dataset}
As shown in Tab.~\ref{tab:robust}, our method achieves leading performance in almost all scenarios. It is worth noting that, since \textit{Android}, \textit{Crab}, and \textit{Yoda} belong to small-scale scenarios with higher observation frequencies, the advantages of our method are less pronounced. However, in large-scale and heavily occluded scenarios such as \textit{Statue}, our method demonstrates significantly greater advantages. As shown in the bottom of Fig.~\ref{statue}, thanks to the observation cognitive correction and the automatic thresholding, our method handles regions with incomplete observations (red circle) and high occlusion more effectively compared to previous SOTA methods.

\subsection{Evaluation on On-the-go Dataset} \label{ongo}
As shown in Tab.~\ref{tab:onthego}, our method maintains leading performance in both high occlusion and low occlusion scenarios. The performance improvement is more significant in high occlusion scenes, which proves that our method can better handle complex, noisy scenes. As shown in the top of  Fig.~\ref{statue}, when only using the default fixed threshold, SplotLessSplats removes too many pixels in low-occlusion scene (\textit{Crab}) and appears too conservative in high-occlusion scene (\textit{Patio high}). On the contrary, our dynamic threshold strategy can effectively handle scenes with different occlusion rates.

\section{Conclusion}
This paper proposes a robust reconstruction framework OCSplasts based on observation completeness and 3DGS, which is used to separate noise elements in scenes containing complex noise and reconstruct clean 3DGS scenes. Our method overcomes the limitation of poor generalization of existing methods. It proposes for the first time a label noise measurement correction theory based on observation completeness, as well as a dynamic anchor threshold calculation method, achieving robust reconstruction of multiple complex scenes without adjusting any hyperparameters. Moreover, the observed completeness obtained through quantification can also be used to determine whether the observations obtained for each region in the scene are sufficient to guide subsequent shooting.

\textbf{Acknowledgements} National Natural Science Foundation of China (Grant nos.62372235,62406069); China Postdoctoral Science Foundation(2024M750425).


%% file: sec/X_suppl.tex
\clearpage
\setcounter{page}{1}
\maketitlesupplementary

\section{Calculation Details of Dynamic Anchor Points $T_b$ and $T_o$}

\textbf{Input.} 
We normalized the input $\mathcal{A}_{h}^{oc}$ and converted it into a histogram $\mathcal{H}$.

\textbf{Details of Calculating $T_o$.}
Firstly, we calculate the probability $P$ of each gray level in the histogram $\mathcal{H}$, the cumulative distribution function $w$, and the cumulative mean $M$:
\begin{equation}
	P(i) = \frac{\mathcal{H}(i)}{N}
\end{equation}
\begin{equation}
	\omega(t) = \sum_{i=0}^{t} P(i)
\end{equation}
\begin{equation}
	M(t) = \sum_{i=0}^{t} i \cdot P(i)
\end{equation}
where $N$ is the total number of pixels in $\mathcal{A}_{h}^{oc}$. $P,\omega$ and $M$ are vectors of length $L$ (in our experiment, $L$=1000).

Next, we calculate the inter-class variance for each grayscale level:
\begin{equation}
	\sigma^2(t) = \frac{[\omega(t) \cdot (M_{\text{global}} - M(t))]^2}{\omega(t) \cdot (1 - \omega(t)) + \epsilon}
	\label{eq28}
\end{equation}
where $M_{\text{global}}=M(L-1)$, $\epsilon=10^{-8}$. 

Finally, we calculate the maximum inter-class variance threshold $T_o$:
\begin{equation}
	t^*=\arg \max _t \sigma^2(t)
\end{equation}
\begin{equation}
	T_o=\frac{t^*}{L}
\end{equation}

\textbf{Details of Calculating $T_b$.}
$T_b$ is the intra-class centroid of the background pixels, which can be calculated by the following formula:
\begin{equation}
	C_0 = \frac{\sum_{i=0}^{t^*} i \cdot P(i)}{\sum_{i=0}^{t^*} P(i)}
\end{equation}
\begin{equation}
	T_b = \frac{C_0}{L}
\end{equation}

\section{Scope of Application}
The OCSplasts in this paper can work well when the noise content of scene labels is above 5\%. When there is less noise, it may cause dynamic threshold failure. We have set up an automatic judgment mechanism in the code; that is, when the max inter-class variance $\sigma^2(t^*)$ in Eq.~\ref{eq28} is greater than $K (K=2000)$, the dynamic threshold is used, and vice versa, a fixed threshold is used.

\section{Stochastic Uncertainty and Threshold Classification}
One notable advantage of using a stochastic uncertainty loss~\cite{martin2021nerf,sabour2023robustnerf} is that it automatically suppresses complex regions to learn by adjusting loss weights, eliminating the need to set thresholds manually. However, because it \textbf{cannot distinguish between truly noisy areas and those that are merely hard to reconstruct, it often yields suboptimal reconstruction quality}. In contrast, threshold classification methods—though lacking generalizability—can produce better reconstruction results when the noise classification is correct. Our proposed dynamic anchor threshold approach circumvents threshold-based methods' poor generalizability while maintaining high reconstruction quality.

\section{More Visualization Results}
\begin{figure*}[!t]
	\centering
	\includegraphics[width=6.8in]{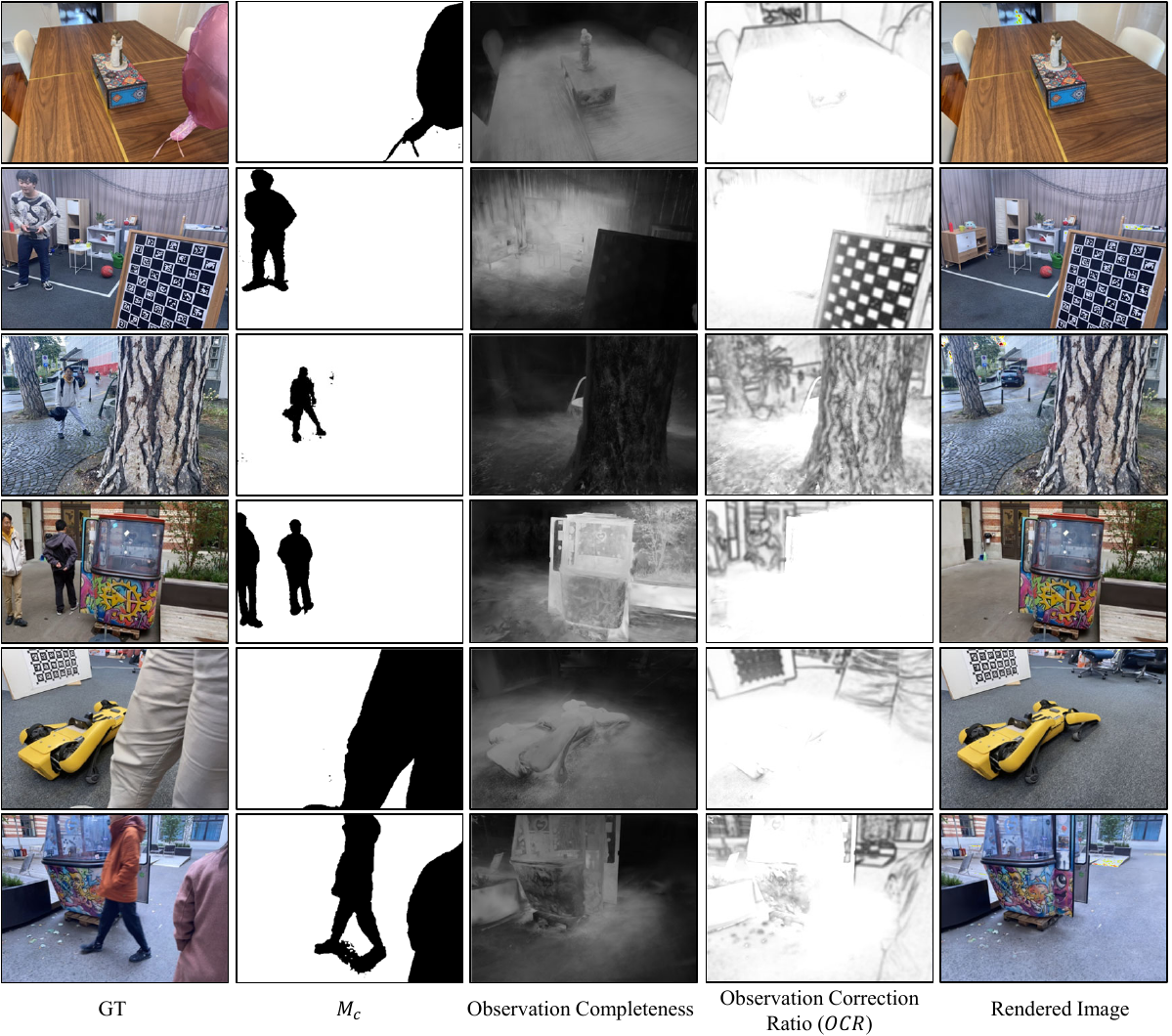}
	\caption{\textbf{Qualitative Results in Indoor and Outdoor Scenes.} OCSplats detect areas with learning difficulties based on observation completeness and use $OCR$ to correct noise assessments, achieving accurate noise segmentation in scenes of different complexities.}
	\label{fig_sup}
\end{figure*}

In Fig.~\ref{fig_sup}, we present more visualized results of OCSplasts on existing datasets. Even if the scene has different proportions of noise and scene characteristics, OCSplasts can still accurately segment foreground noise and reconstruct clear static backgrounds.
\begin{figure}[!t]
	\centering
	\includegraphics[width=3.4in]{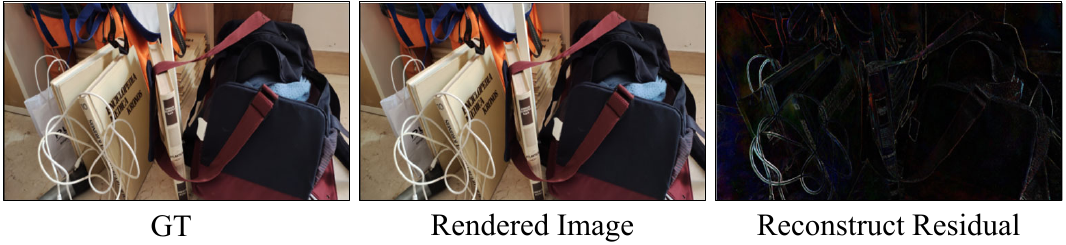}
	\caption{\textbf{Pose Error.} Even in completely stationary scenes, camera pose errors can cause edge position reconstruction to fail.}
	\label{fig_inherent}
\end{figure}

\section{Limitations}
Currently, most reconstruction methods rely on the pose provided by COLMAP~\cite{schoenberger2016sfm,schoenberger2016mvs}, but we have found that the COLMAP pose may have some errors. This kind of pose error is often reflected in the texture edges of the object, as shown in Fig.\ref{fig_inherent}. Even in entirely stationary scenes, the reconstruction interference caused by slight pose misalignment is inevitable. Similarly, errors at these edges are objectively present in reconstructing scenes containing dynamic objects, so these inherent edge noises have been interfering with the noise measurement. In our future work, we will attempt to evaluate camera pose errors and quantify these inherent edge interferences to eliminate their impact on noise label separation.